\documentclass[10pt, conference, compsocconf]{IEEEtran}
\IEEEoverridecommandlockouts

\usepackage{cite}
\usepackage{amsmath,amssymb,amsfonts}
\usepackage{algorithmic}
\usepackage{graphicx}
\usepackage{textcomp}
\usepackage{xcolor}
\usepackage{xspace}
\usepackage{comment}
\usepackage{caption}
\usepackage{subcaption}

\usepackage{hyperref}
\usepackage[numbers]{natbib}
\usepackage{multirow}

\newcommand{\ie}{{i.e.,}\xspace}
\newcommand{\eg}{{e.g.,}\xspace}
\newcommand{\namemeaning}{Our system's name is inspired by the Greek God who presides over passages, doors, gates, and endings. We aim for our system {\sc Janus\xspace} to be able to carve out the correct transition to right-sized algorithms for the world to get maximum performance per dollar cost, since he looks to the future and to the past.\xspace}
\newcommand{\name}{{\sc Janus}\xspace}

\begin{document}
\pagestyle{plain}

\title{\name: Benchmarking Commercial and Open-Source Cloud and Edge Platforms for Object and Anomaly Detection Workloads}
\author{Karthick Shankar, Pengcheng Wang, Ran Xu, Ashraf Mahgoub, Somali Chaterji\\\textit{Purdue University, West Lafayette, IN, USA}}

\maketitle
\begin{abstract}
With diverse IoT workloads, placing compute and analytics close to where data is collected is becoming increasingly important. We seek to understand what is the performance and the cost implication of running analytics on IoT data at the various available platforms. These workloads can be compute-light, such as outlier detection on sensor data, or compute-intensive, such as object detection from video feeds obtained from drones. In our paper, \name, we profile the performance/\$ and the compute versus communication cost for a compute-light IoT workload and a compute-intensive IoT workload. In addition, we also look at the pros and cons of some of the proprietary deep-learning object detection packages, such as Amazon Rekognition, Google Vision, and Azure Cognitive Services, to contrast with open-source and tunable solutions, such as Faster R-CNN (FRCNN). 
We find that AWS IoT Greengrass  delivers  at  least  2X  lower  latency  and  1.25X  lower cost compared to all other cloud platforms for the compute-light outlier detection workload.
For the compute-intensive streaming video analytics task, an open-source solution to object detection running on cloud VMs saves on dollar costs compared to proprietary solutions provided by Amazon, Microsoft, and Google, but loses out on latency (up to 6X). If it runs on a low-powered edge device, the latency is up to 49X lower.
\end{abstract}

\begin{IEEEkeywords}
sensor data outlier detection, object detection, AWS EC2, AWS IoT Greengrass, AWS Lambda
\end{IEEEkeywords}

\IEEEpeerreviewmaketitle

\section{Introduction}
Cloud computing (a.k.a \textit{Infrastructure-as-a-service}) is becoming the main execution environment for many users due to its ease of management, scalability, and fault-tolerance. By removing the need for hardware and cluster management, users can now focus on their application needs and have a finer-granularity pricing model for their resource usage, even more so with the advent of serverless computing where the billing is compute-driven rather than time-driven~\cite{jonas2019cloud, akkus2018sand}. For example,  Amazon AWS provides a virtualized server-based computing service \textit{Amazon Elastic Compute Cloud or Amazon EC2}.
Amazon also provides a serverless computing service, \textit{AWS Lambda}, which allows users to execute their code without having to provision or manage servers. Users essentially pay for the exact amount of allocated resources and the compute time (in 100ms increments) with no charge for idle time. With serverless computing, applications can automatically scale instantaneously by running code in response to events or triggers. \textit{Amazon IoT Greengrass} further extends the AWS infrastructure to the edge and provides lower latency computation through running code on the same IoT device(s) that collects the data. \\
The three services vary in their strengths and deciding which service to use for a given workload is not trivial for several reasons: (1) Users have different \$ budget and performance requirements. (2) Real-world workload characteristics often vary over time~\cite{Rafiki2017, Sophia2019, optimuscloud2020, xu2020approximate}, \eg streaming video analytics can be compute-intensive in case of fast-changing scenes and becomes lighter for relatively static scenes. (3)  The \$ cost with each service varies with time and geographical regions. 
(4) Different services have different types of limitations, which may make it impossible to run a particular application on some service. For example, in AWS Lambda, a function has a time limit of 15 minutes, which makes running complex, stateful algorithms difficult. Also, picking exact configurations for the instances that run the serverless code is not possible (users can only specify the memory requirements, and other resources are scaled accordingly). Therefore, quantitative 
evaluation is needed with representative applications in order to identify the appropriate computing framework for the applications, to explore the trade-off between accuracy, performance, \$ cost, and configurability.  

In \name\footnote{\namemeaning}, we compare several cloud computing services for two representative IoT applications that vary in complexity. The first is a simple outlier detection application for sensor data, and the second, is a complex object detection application on streaming video. Both of these algorithms are ubiquitous in IoT, rely on online data streaming, and provide contrasting bandwidth requirements and algorithmic processing capabilities~\cite{yu2017recursive, roady2020stream}. Therefore, we select these two applications as representative workloads for compute-light vs. compute-intensive IoT applications. 
IoT devices are used for simple data acquisition in many scenarios, like in farms~\cite{chaterji2020artificial, jiang2020hybrid} and for self-driving cars~\cite{chaterji2019resilient, bagchi2019grand}. Since the volume of data acquired through these sensors is high, it is often run through an outlier detection program to ensure proper analysis of data and to discover faulty sensors. For instance, in the case of farm sensor data, a farmer would want to know the real-time temperature and humidity of the farm, and any delayed intervention may lead to losses in yield and consequent financial losses. For the object detection workload, many works like~\cite{LingHu2018IAOD} use different object detection algorithms on IoT devices for a variety of situations like gaze detection and surveillance in smart cities. Real-time object detection is essential to security-critical or latency-critical scenarios like self-driving cars, or less critically, yet increasingly prevalent, in large crowds at mass entertainment events. Thus, these two workloads represent popular options for IoT applications while showing variety in compute requirements. 

We perform benchmarking experiments with these applications on two different platform types---{\bf edge computing} and {\bf cloud computing} platforms~\cite{suryavansh2019tango}. Within the edge computing platform type, we explore the commercial offerings, AWS Greengrass and Google IoT Edge and two different types of compute nodes, a Raspberry Pi and a Docker container, the latter to emulate more resource-rich devices like the Nvidia Jetson series. Within the cloud computing platform type, we conduct experiments on three commercial offerings, Amazon EC2, Google Compute, and Microsoft Azure Virtual Machine. 
Our goal is to aid in selection of the best platform for each target application. In addition, given the huge increase in demand for streaming video analytics (such as object detection), we profile three leading commercial offerings, Amazon Rekognition, Google Vision, and Azure Cognitive Services, to benchmark against a popular open-source region-based CNN using attention mechanisms called Faster R-CNN (FRCNN)~\cite{ren2015faster}. We also use FRCNN to show possible trade-offs between latency and accuracy that can impact the end-to-end \$ cost. We use FRCNN, as opposed to other popular object detection algorithms \eg YOLO and SSD, since it has a higher accuracy with classification and bounding box-regression in consecutive stages, at the expense of computational complexity (useful to showcase \name's compute-intensive use case). 

In this paper, we ask three questions vis-\`a-vis the computing platforms and software packages described above.
\begin{enumerate}
    \item What platform to run an IoT workload on, on the cloud and on the edge, respectively for a compute-intensive and for a compute-light workload? 
    \item What is the latency and \$ cost of running on each platform?
    \item What is the advantage of using an open-source object detection framework on a cloud-based virtual machine over using the commercial offerings? 
\end{enumerate}

Following are the chief insights that come out of \name.
\begin{enumerate}
    \item Our benchmarking of the compute-light IoT workload reveals that AWS Greengrass delivers at least 2X lower latency and 1.25X lower cost compared to all other platforms for this workload (Tables~\ref{tab:aws_greengrass_pi_outlier_detect} and \ref{tab:aws_Google_IoT_outlier_detect}).
    \item Our benchmarking of the compute-intensive object detection algorithm on streaming video on Amazon Rekognition, Google Vision, and Faster R-CNN (on Amazon EC2) reveals that Faster R-CNN is 12.8X to 21.0X cheaper than Amazon Rekognition and Google Vision solutions but is also much slower than the others (Table~\ref{tab:rekog_vision_object_detection}). Also, we propose a novel approximation of Faster R-CNN and show that we can flexibly navigate the space of latency versus accuracy. In contrast, for the commercial offerings, no such tradeoff is possible. Among the commercial offerings, Google Vision is faster but less performant in latency/\$ terms than Amazon Rekognition (55\% less) and Azure Cognitive Services (11\% less).
    \item To delve deeper into the open source Faster R-CNN, we execute it on the three commercial cloud platforms---Amazon EC2, Microsoft Azure, and Google Compute. We find that we can execute more frames per \$ on Google Compute than EC2 (94\%) and Azure (153\%). We also see that one approximation knob (the number of region proposals in FRCNN) has significant effect on the running time---the running time is reduced by 57.3\% when approximating aggressively compared to the default parameter value while the accuracy is reduced by only 9\% (Table~\ref{tab:rekog_vision_object_detection}). This tunability also has the additional interpretability benefit, which is helpful in several domains~\cite{kim2016opening}.
\end{enumerate}
\section{Background}
Here we give a brief description of the foundational platform and the different commercial edge computing platforms and the vision services that we benchmark in this paper. We also describe the open-source object detection software package, Faster R-CNN. Since the commercial cloud computing platforms we consider here are so commonplace, we omit their background information. 

\textbf{Edge computing} is the practice of placing computing
resources at the edges of the Internet in close
proximity to devices and information sources. This,
much like a cache on a CPU, increases bandwidth
and reduces latency for applications but at a potential
cost of dependability and capacity~\cite{bagchi2019dependability}. This is because
these edge devices are often not as well maintained,
dependable, powerful, or robust as centralized server class cloud resources.

The edge paradigm supports the large scale IoT devices, where real-time data is generated based on interactions with the local environment. This complements more heavy-duty processing and analytics occurring at the cloud level. This structure serves as the backbone for applications, such as augmented reality and home automation, which utilize complex information processing to analyze the local environment to support decision making. In the IoT domain, functional inputs and outputs are physically tied to geographically distributed sensors and actuators. If this data is processed in a central location, immense pressure will be placed on ``last mile'' networks, and cloud-leveraged IoT deployments will become impractical.
\begin{figure}[ht]
    \centering
    \includegraphics[width=\linewidth]{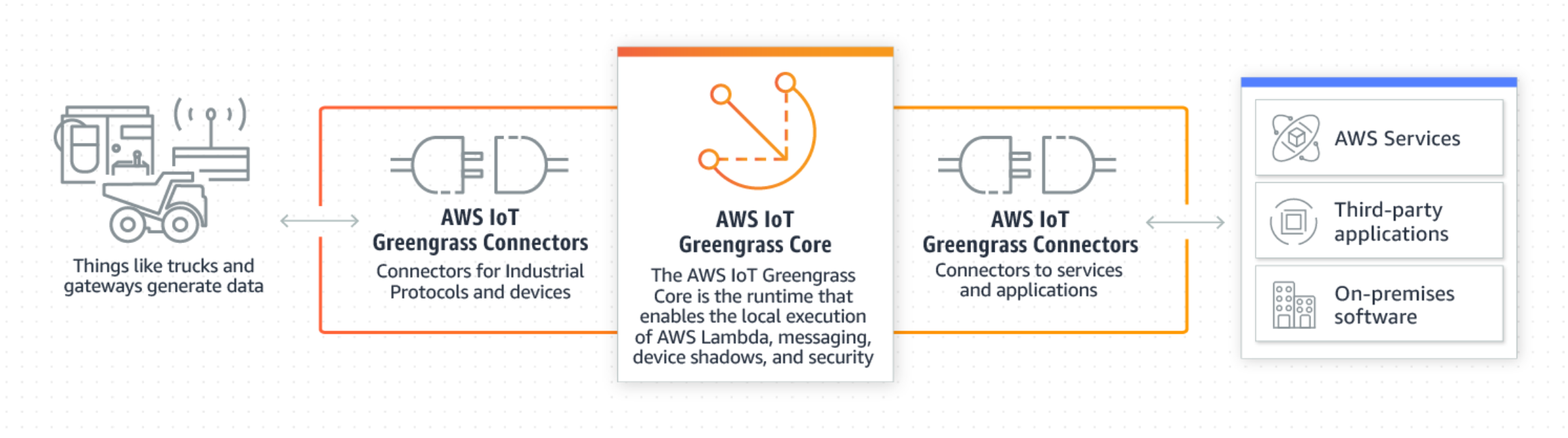}
    \caption{AWS Greengrass Architecture\protect~\cite{gg-arch}}
    \label{fig:gg-arch}
\end{figure}
    
\begin{figure}[ht]
    \centering
    \includegraphics[width=\linewidth]{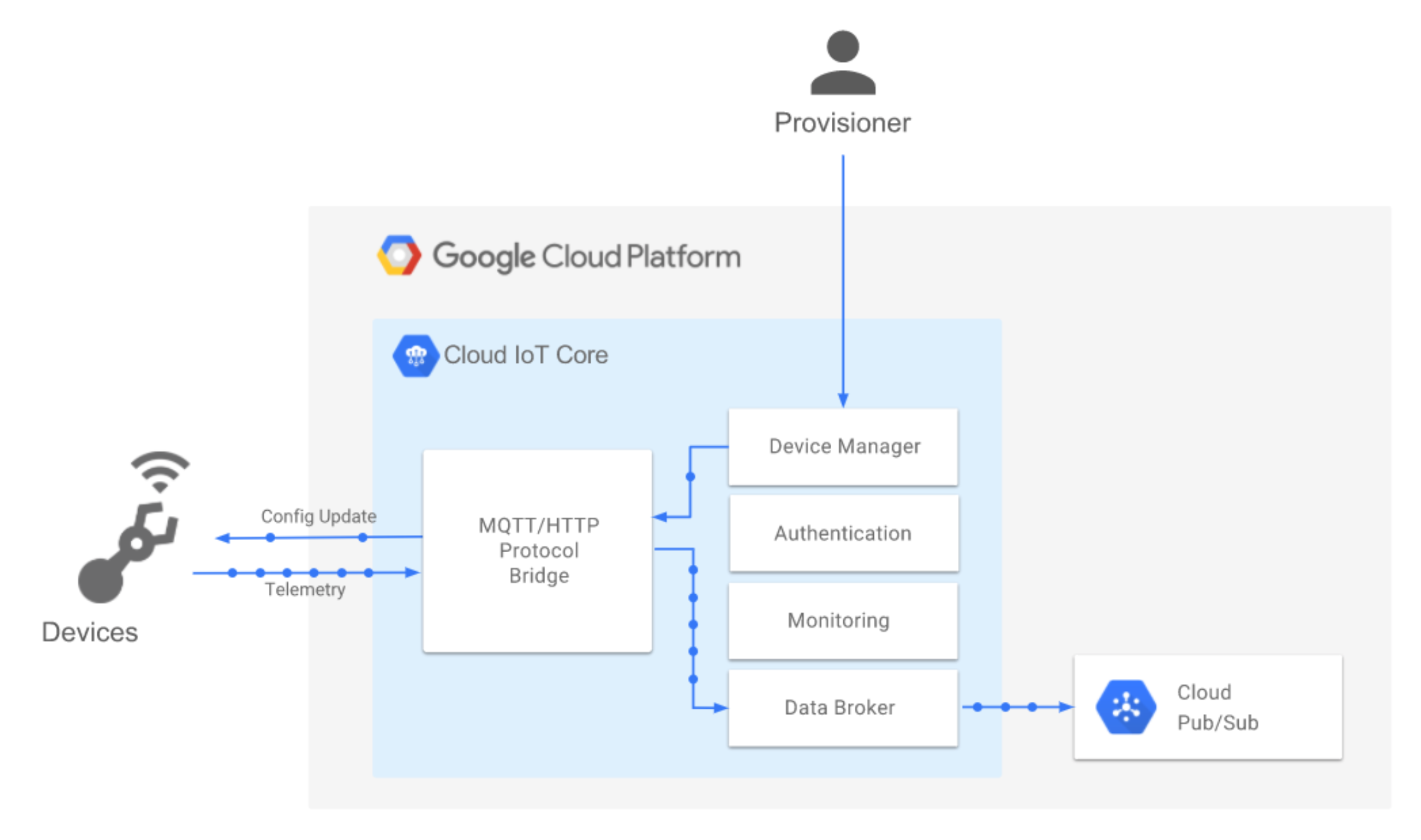}
    
    \caption{Google IoT Core Architecture\protect~\cite{google-iot-core}}
    \label{fig:iot-core-arch}
\end{figure}
\textbf{AWS Greengrass} is a service offered by Amazon (initially as an IoT gateway, now morphed into an edge computing service) that enables data management, durable storage, cloud analytics, and local computing capabilities for connected edge devices. Notice that Greengrass does not provide any compute power itself and should be looked upon more as an orchestrator among devices that are outside of the AWS framework and provided by the user. Connected devices can run AWS Lambda functions or Docker containers, while data and control flow to these devices through the Greengrass framework. Subsets of the information generated at the edge can be communicated back to the AWS Cloud. Greengrass also keeps devices' data in sync, and securely communicates with other devices, even when not connected to the Internet. This means that Greengrass-connected IoT devices can still respond quickly to local triggers, interact with local resources and minimize the costs associated with transmitting IoT data to the cloud. Its architecture is shown in Figure~\ref{fig:gg-arch}\\
\textbf{Cloud IoT Core} is a service offered by Google that allows secure connectivity, management, and ingestion of data from millions of globally dispersed devices for operational efficiency.~\cite{google-iot-core}. Cloud IoT Core runs on Google’s serverless infrastructure, which adaptively scales horizontally in response to real-time events. Like in the Greengrass case, the user has to provide the device on which the computation of the Cloud IoT Core will run. Cloud IoT core supports the standard MQTT (Message Queue Telemetry Transport, essentially the messaging protocol for IoT) and HTTP protocols, making it easier for devices to be registered to Cloud IoT Core. Its architecture is shown in Figure~\ref{fig:iot-core-arch}\\
\textbf{AWS Greengrass versus Cloud IoT Core}: 
For AWS Greengrass, \textit{AWS IoT Greengrass Core} provides local services (compute, messaging, state, security), and communicates locally with devices that run the AWS IoT Device SDK~\cite{gg-faqs}. For Google Cloud, the IoT Core provides the services to communicate with the various IoT devices that have been registered to it.
As such, a key difference between the two is that Greengrass works with devices by running AWS Lambda functions and communicating through the SDK while Google IoT Core uses the standard MQTT (a machine-to-machine telemetry protocol) or HTTP protocol for communications. 
AWS Greengrass and Google IoT Core both present a gateway between the edge IoT devices and more powerful cloud services. They act as connectors to move data between the edge and the cloud. With AWS Greengrass, Lambda functions are run between the edge machines through the AWS Greengrass SDK while with Google IoT core, MQTT commands are used.

\textbf{Pricing differences}: In Amazon's \textit{AWS IoT Greengrass}, payments are structured per the number of \textit{AWS IoT Greengrass Core} devices that are connected and interact with the \textit{AWS Cloud} in a given month. This price depends on the region that is configured with \textit{Greengrass} and ranges from \$0.16--\$0.18 per month per \textit{IoT Greengrass Core} device. There is no additional cost for the number of \textit{AWS IoT SDK} enabled devices locally connected to any \textit{IoT Greengrass Core} \cite{AWS-GG-Pricing}. However, there can be additional charges with \textit{AWS IoT Greengrass} if data transfer or any other AWS service is involved in the application, the pricing of which depends on the service used. Amazon S3 is commonly used when there is a large quantity of data that needs to be stored and processed elsewhere. In compute-light applications where the edge device does some processing and sends just a simple result, \eg outlier detection where the number of outliers are sent as a result, S3 is not needed. In contrast, for compute-intensive applications that involve images or large data sets, S3 can be used to store the entire data set while the edge devices download and process small chunks. In our evaluations, we do all the processing in the device with no extra storage overhead for the outlier detection. For object detection, we store the videos on {\bf the sensor device} and upload to the services API for processing, thus incurring no cost for cloud storage.
In contrast, Google's \textit{IoT Core} pricing is tiered according to the data volume used in a calendar month~\cite{iot-pricing}. 
This volume is based on all data exchanges between the devices connected to \textit{Google IoT Core}. Cloud IoT Core is priced per MB of data exchanged by IoT devices with the service after a 250MB free tier. After that, in the initial range, the price is \$0.0045 per MB and goes down to \$0.00045 per MB at the high end of the range, beyond a certain threshold of data usage. 

\textbf{Amazon Rekognition, Google Vision, and Azure Vision Services for our object detection application}:
Amazon Rekognition provides an API for analyzing images (Amazon Rekognition Image) which we use for streaming video analysis. Rekognition uses deep learning algorithms with SDKs for many programming languages and requires no machine learning expertise. With Amazon Rekognition, one can identify objects, people, text, scenes, and activities in images and videos, as well as flag inappropriate content. Amazon Rekognition also provides facial analysis and facial search capabilities that one can use to detect, analyze, and compare faces for user verification, people counting, and public safety use cases~\cite{aws-rekog}. It provides an easy-to-use API that returns the results of computation but without being able to control the backend of the computation. In addition, although we are able to control the AWS availability zone for the VM selection (for latency considerations), the selection may be too coarse for applications with strict low latency requirements such as in autonomous driving.

Google Vision and Azure Cognitive Services are similar to Amazon Rekognition in that they are also an image analysis service that offer powerful pre-trained machine learning models through REST and RPC APIs. Google Vision can detect objects and faces, read printed and handwritten text, and build valuable metadata into an image catalog~\cite{gvision}. Azure Cognitive Services also provides form and ink recognition to analyze written documents and handwriting. 


The use cases for these services greatly depend on the application and the problem that needs to be solved. Amazon Rekognition platforms have support for popular tasks like object detection, celebrity recognition, face recognition, content moderation, and text detection. Amazon Rekognition also offers the Pathing option that allows users to run videos through the service and see the paths that the people in the video take~\cite{aws-rekog}. Google Vision on the other hand offers product search options to scan a product and quickly find similar listings~\cite{gvision}. Azure Vision also has face recognition technology similar to Amazon Rekognition.


{\textbf{Customized vision applications versus commercial offerings}}---\textbf{engineering challenges and solutions for the data engineer}: The APIs in all three platforms ease the process of prototyping a computer vision application. However, we also notice that the developers are not able to specify the backend compute infrastructure on which it will run. For example, we are not able to leverage our edge device to force the services to run next to our data storage. The developers are also not able to use their own model or select a model to run or tune the configuration knobs of the model for desired accuracy/runtime/energy specifications. 
Considering the two challenges above, data engineers can leverage their own computer vision applications on AWS EC2, AWS Lambda, or AWS IoT Greengrass. This is where an open-source software package like Faster R-CNN comes into play. 

{\textbf{Faster R-CNN (FRCNN)}}~\cite{ren2015faster}: FRCNN is a state-of-the-art object detection algorithm based on using region proposal networks to hypothesize object locations. Thus it speeds up upon its earlier versions---both R-CNN and Fast R-CNN use selective search to find region proposals~\cite{girshick2015fast, girshick2014rich}. Selective search is slow, affecting the performance of the network. In contrast, FRCNN uses a separate network to predict the region proposals (\textit{Region Proposal Networks}, RPNs). RPNs are designed to efficiently predict region proposals (with a high recall), with a wide range of sizes and aspect ratios, by using novel ``anchor'' boxes to serve as references at differential scales and aspect ratios. Region proposals are then reshaped using a region of interest (RoI) pooling layer, which essentially uses inputs of non-uniform sizes to obtain fixed-size feature maps. This RoI pooling layer then classifies the image within the proposed region and further refines the bounding boxes (regressor).
We add to FRCNN, different {\em levels of approximation} that can be tuned at runtime, for different points in the latency vs. accuracy space. An easy-to-adjust approximation parameter is the number of proposals that an RPN generates, which by convention is set to be the largest possible number of objects in the image. Since the classifier and bounding box regressor are region-wise, a smaller number of proposals reduces the execution cost, at the risk of reduced accuracy when a large number of objects exist in the image. This notion of context-aware approximation has been introduced in some domains, like genomics~\cite{koo2018tiresias}, and closer to our application context, streaming video processing~\cite{xu2018videochef, xu2019approxnet}. Here our objective is to expose this novel approximation knob and to show that this kind of configurability is present only in the open source options. 



\section{Experimental Setup}
\label{sec:Experimental_Setup}
Here we describe the benchmark data sets used in the study, the workload analysis performed (Outlier detection vs. Object detection), and the experimental setup for the different platforms used in our study. These platforms include: 
\begin{itemize}
  \item Edge Devices
  \begin{enumerate}
      \item Raspberry Pi 4 model B
      \item Emulated Edge device (using Docker Containers)
  \end{enumerate}
  \item Cloud Platforms
  \begin{enumerate}
      \item Amazon EC2
      \item Google Compute
      \item Microsoft Azure Virtual Machine
  \end{enumerate}
  \item IoT Managers
  \begin{enumerate}
      \item AWS Greengrass
      \item Google IoT Core
  \end{enumerate}
  \item Other Commercial Offerings
  \begin{enumerate}
      \item Amazon AWS Lambda (serverless functions)
      \item Amazon Rekognition
      \item Google Vision
      \item Microsoft Azure Cognitive Services
  \end{enumerate}
\end{itemize}
We do an exhaustive assessment of these cloud, edge, and IoT orchestration platforms to evaluate the efficacy of different vendors' hardware platforms, architectures, or networking protocols. Figures 3 and 4 show the setups for both workloads.

\subsection{Outlier-Detection data description}
The data we used for this benchmarking contains 21k points collected from February to October 2019 using temperature and humidity sensors deployed in sensorized farms and manufacturing units on Purdue University's campus. We apply extreme value analysis (EVA), which is a popular and simple statistical analysis to identify outliers in the data. In this analysis, we fit a Gaussian distribution to the data and use standard statistical outlier detection. Under the Gaussian distribution assumption, we expect that 68\% of the data points will be within one \textit{Standard deviation} from the \textit{Mean} and 95\% to be within two \textit{Standard deviations} from the \textit{Mean}. We use this distance from the \textit{Mean} as our outliers' cut-off threshold. Tables~\ref{tab:aws_lambda_outlier_detect} -- \ref{tab:microsoft_azure_outlier_detect} show the number of outliers in both temperature and humidity readings with varying cut-off thresholds. Figure~\ref{fig:temperature_humidity_variation} shows the temperature and humidity variation of a single device.

\subsection{Object-Detection data description}
For the compute-intensive workload analysis, we use video data from the ImageNet Large Scale Visual Recognition Challenge (ILSVRC) 2015~\cite{ILSVRC15} for video object detection. The evaluation data set contains 555 video snippets with labels of 30 object categories as ground truth. These videos are good representations of real captured videos from surveillance cameras or drone cameras. We perform object detection on these videos---\ie classify rectangular regions on each frame into one from the 30 object categories. We assume the video data is stored on the sensor device and the processing (object detection) is done using different proprietary algorithms through commercial offerings or using the variants of the Faster R-CNN model on a cloud virtual machine, which is a widely used custom whitebox model.

\subsection{Infrastructure setup}
In our experiments, we use a Raspberry-Pi 4 model B as our edge device. This model has a Broadcom BCM2711, Quad core Cortex-A72 (ARM v8) 64-bit SoC @ 1.5GHz, 4GB LPDDR4-3200 SDRAM, which is one of the most popular edge devices for developers. We also use Docker containers as additional emulated edge devices with higher compute capability, similar to higher powered devices like the NVIDIA Jetson. We use this strategy to be able to adaptively control the edge specs while not needing additional hardware. This, in addition to the real edge devices, this gives us the opportunity to be able to have a platform for trying out different edge specifications.

We use 1 CPU and 1 GB RAM as our Docker containers' specification for sensor and edge devices. Our server, coming with a six-core Intel Xeon CPU E5-2440 clocked at 2.40GHz and 48GB RAM, is powerful enough to simulate multiple sensor and edge devices. Ismal \textit{et al.} in~\cite{dockeredge} show that Docker containers provide fast deployment, small footprints, elasticity, and good performance, which enable them to simulate edge devices. Furthermore, Docker images are small and lightweight, 
making the CPU, memory, storage, and network performance similar to physical edge devices~\cite{dockerperf}. 

We also use IoT orchestrators (AWS Greengrass and Google IoT core) for both the outlier detection and object detection experiments on the Raspberry Pi and Docker container (emulated edge device). This is to account for scalability if processing is done on multiple edge devices. For the cloud platform experiments, this is not necessary since the data is being sent to a central storage location and not individually processed by each device.
Our cloud infrastructure is as follows for the different platforms:
\begin{enumerate}
    \item \textit{Amazon EC2}: c5.large (2 vCPUs, 4 GiB memory)
    \item \textit{Google Compute}: e2-standard (2 vCPUs, 8 GiB memory)
    \item \textit{Microsoft Azure Virtual Machine}: Standard F2s\_v2 (2 vCPUs, 4 GiB memory)
\end{enumerate}

In the cases where we use commercial offerings that have more sophisticated object detection algorithms, we are forced to use the APIs provided by the vendors without the ability to control the backend device or any parameters.
\section{Evaluation}

\begin{figure*}[ht]
    \centering
    \includegraphics[width=0.8\linewidth]{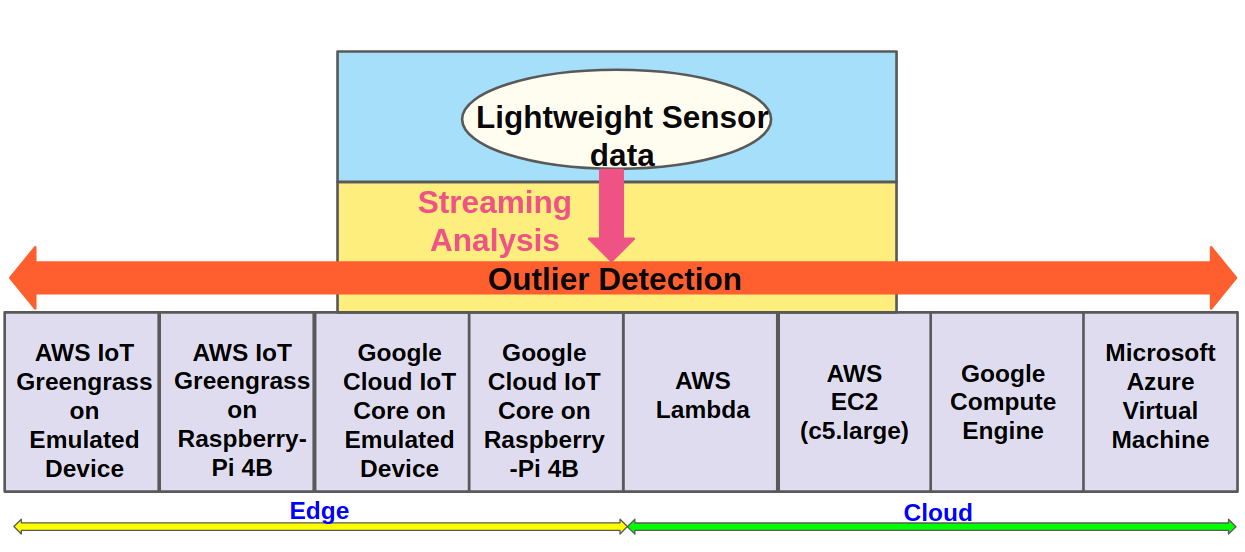}
    \label{fig:overall_outlier}
    \caption{Overall Setup for Outlier Detection Evaluation. 
    Data is collected from the IoT sensor devices (temperature and humidity) and stored/streamed to the appropriate storage option for different platforms. For example, we store the data in Amazon S3, which is a remote storage, for cases where the processing is on the cloud or on AWS Lambda. We store the data into the device's local storage when the processing is on the edge device. We then perform the same analysis across all the platforms and report the latency and \$ cost for each service. By tuning the threshold for outlier detection, we can get different proportions of outliers}
    \centering
\end{figure*}

\begin{figure}[ht]
    \centering
    \includegraphics[width=\linewidth]{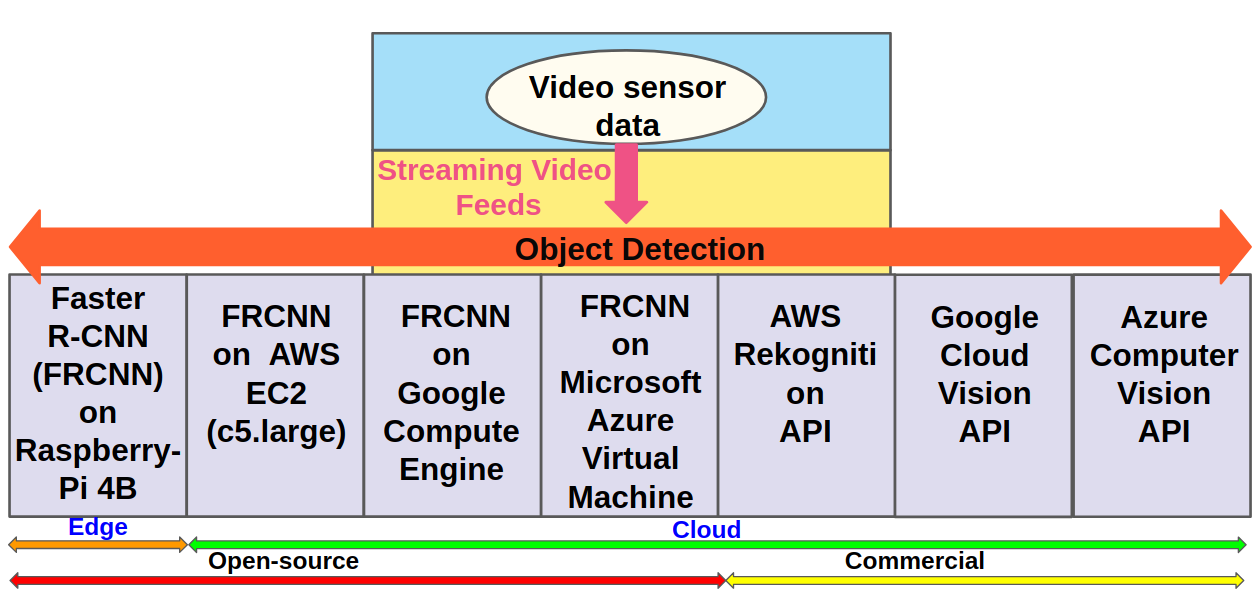}
    \label{fig:overall_object}
    \caption{Overall Setup for Object Detection Evaluation. 
    The video data set is ILSVRC VID 2015 and is stored on the embedded device (Raspberry Pi or an Emulated device using a Docker Container). It is directly uploaded using the vendor's API if needed, and uploaded to AWS S3 for processing on cloud platforms. Our custom algorithm Faster R-CNN (FRCNN), where the customization knob is the number of proposals, is used for comparison with three commercial services in their corresponding platforms. We report execution time/frame and \$ cost for each service on each platform.}
\vspace{-2mm}
\end{figure}

\subsection{Data Preprocessing}
For the outlier-detection application, we use temperature and humidity data collected from 26 WHIN-IoT devices. We divide the data into monthly segments. Next, we assume that the temperature and humidity will be normally distributed (Gaussian), and we compute the mean and standard deviation of monthly measurements to identify outliers. Here, the goal and motivation of outlier detection 
is to track if the sensors are malfunctioning.

\begin{figure}
\begin{subfigure}{\linewidth}
  \centering
  \includegraphics[width=\linewidth]{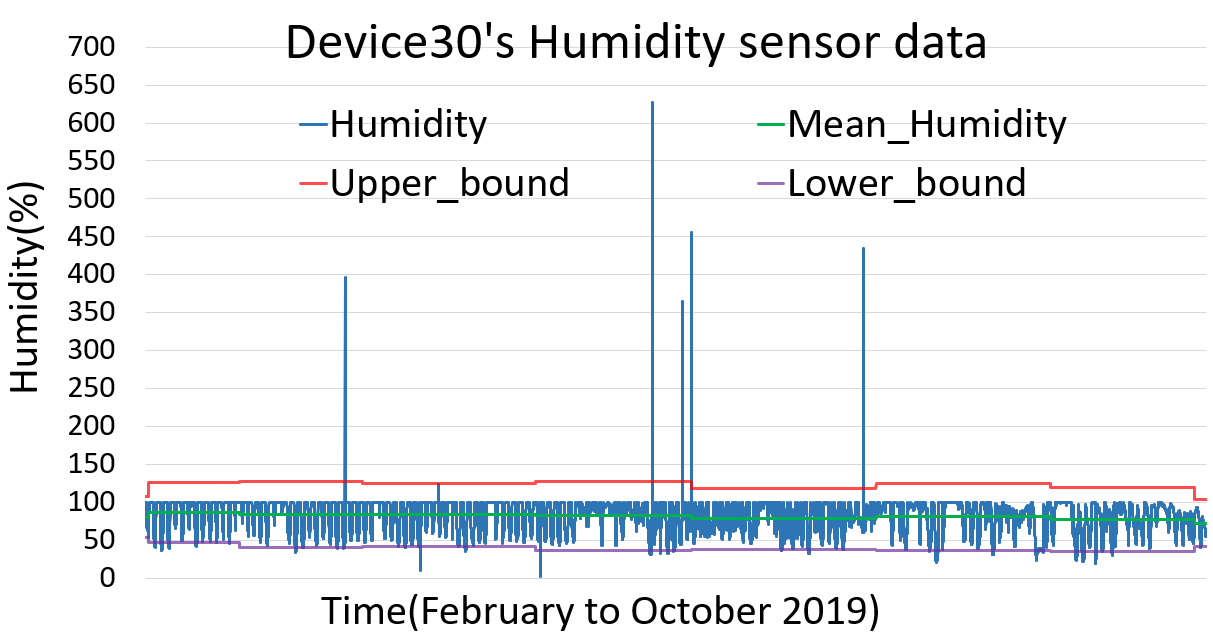}  
  \caption{Humidity Variation for Device 30}
  \label{fig:sub-first}
\end{subfigure}

\begin{subfigure}{\linewidth}
  \centering
  \includegraphics[width=\linewidth]{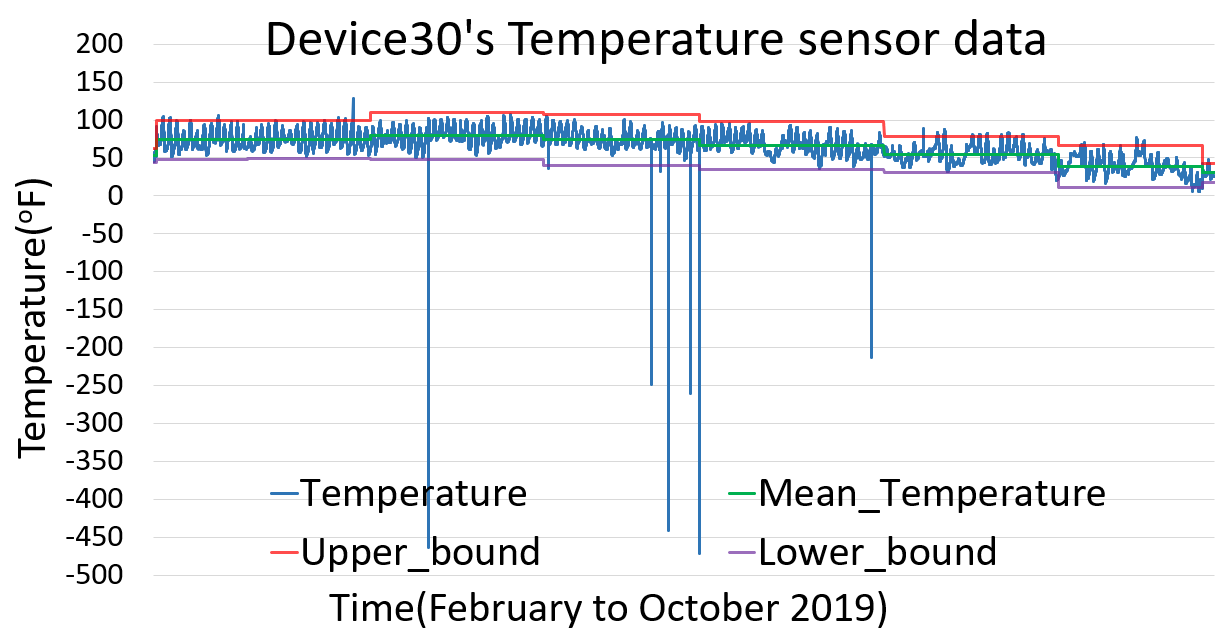}  
  \caption{Temperature Variation for Device 30}
  \label{fig:sub-second}
\end{subfigure}
\caption{Device 30's temperature and humidity variation and outlier detection thresholds}
\label{fig:temperature_humidity_variation}
\end{figure}


For the object-detection application, we download the widely used ILSVRC 2015 video dataset~\cite{ILSVRC2015-video} to the sensor device as a stand-in for the captured videos on the sensors. We use Amazon Rekognition, Google Vision, Azure Cognitive Services, and our custom ``service'' (Faster R-CNN on cloud VMs) for object detection on these videos. The video data is processed in a streaming manner with each frame being processed separately.

Overall, we wanted to benchmark using the ubiquitous sensor data sets that are generated in different urban and rural IoT settings such as smart factories~\cite{thomas2018minerva} or connected farms~\cite{chaterji2020artificial}. We do analysis on different platforms offered by Amazon, Microsoft and Google. These platforms provide different virtual machine specifications and price structures.

\begin{table*}[th]
\setlength{\abovecaptionskip}{0pt}
\begin{center}
\scriptsize
\def\tablename{Table}
\caption{\upshape{Performance and cost metrics for running our compute-light operation, specifically, outlier detection on AWS-Lambda.}}
\begin{tabular}{|c|c|c|p{0.6in}|c|c|c|c|} 
 \hline
 Metric & Temperature outliers (\%) & Humidity outliers (\%) & Data-Passing Duration (ms) & Duration (ms) &  Billed Duration (ms) & Memory Size (MB) & \$ cost\\ 
  \hline
 $\mu \pm 1 \times \sigma$ & 5,978 (28.165\%) & 5,706 (26.883\%) & 549.845 & 1,045.64 & 1,100 & 92 & \$0.000004587 \\
  \hline
 $\mu\pm2\times\sigma$ &	446 (2.101\%) &	561 (2.643\%) & 605.557 & 1,104.7 &	1,200 &	 92 & \$ 0.000005004 \\
  \hline
 $\mu\pm3\times\sigma$ &	6 (0.028\%) & 5 (0.024\%) & 545.787 &	1,063.51 &	1,100 &	 93 & \$0.000004587 \\
 \hline
\end{tabular}
\label{tab:aws_lambda_outlier_detect}
\end{center}
\vspace{-5.5mm}
\end{table*}

\subsection{Experiments and Results (Outlier Detection)}

\begin{table*}
\setlength{\abovecaptionskip}{0pt}
\begin{center}
\scriptsize
\def\tablename{Table}
\caption{\upshape{Performance and cost metrics for performing outlier detection on an emulated edge device with AWS Greengrass as the IoT manager. Greengrass provides a flat monthly billing.}}
\begin{tabular}{ |c|c|c|p{0.6in}|c|c|c|c| } 
 \hline
 Metric & Temperature outliers (\%) & Humidity outliers (\%) & Data-Passing Duration (ms) & Duration (ms) &  Billed Duration (ms) & Memory Size (MB) & \$ cost\\ 
  \hline
 $\mu\pm1\times\sigma$ &	5,978 (28.165\%) &	5,706 (26.883\%) & - &	52.099 & - &	92 & \$ 0.0000037\\
  \hline
 $\mu\pm2\times\sigma$ &	446 (2.101\%) &	561 (2.643\%) & -  & 54.577 & - &	 92 & \$ 0.0000037\\
  \hline
 $\mu\pm3\times\sigma$ &	6 (0.028\%) &	5 (0.024\%) & - & 52.490 & - &	 93 & \$ 0.0000037\\
 \hline
\end{tabular}
\label{tab:aws_green_grass_outlier_detect}
\end{center}
\end{table*}

\begin{table*}
\setlength{\abovecaptionskip}{0pt}
\begin{center}
\scriptsize
\def\tablename{Table}
    \caption{\upshape{Performance and cost metrics for performing outlier detection on Raspberry-Pi 4B with AWS Greengrass as the IoT manager. }}
\begin{tabular}{ |c|c|c|p{0.6in}|c|c|c|c| } 
 \hline
 Metric & Temperature outliers(\%) & Humidity outliers(\%) & Data-Passing Duration (ms) & Duration (ms) &  Billed Duration (ms) & Memory Size (MB) & \$ cost\\ 
  \hline
 $\mu\pm1\times\sigma$ &	5,978 (28.165\%) &	5,706 (26.883\%) & - &	178.27 & - &	12 & \$0.0000037\\
  \hline
 $\mu\pm2\times\sigma$ &	446 (2.101\%) &	561 (2.643\%)	 & -  & 172.09 & - &	 12 & \$0.0000037\\
  \hline
 $\mu\pm3\times\sigma$ &	6 (0.028\%) &	5 (0.024\%) & - &	163.72 & - &	 12 & \$0.0000037\\
 \hline
\end{tabular}
\label{tab:aws_greengrass_pi_outlier_detect}
\end{center}
\end{table*}

\begin{table*}
\setlength{\abovecaptionskip}{0pt}
\begin{center}
\scriptsize
\def\tablename{Table}
\caption{\upshape{Performance and cost metrics for outlier detection on  an emulated edge device with Google IoT. Unlike AWS Greengrass, Google IoT charges for only the data transfer between edge devices or between the edge and cloud. We estimate the price and neglect the initial free data volume per account.}}
\begin{tabular}{ |c|c|c|p{0.6in}|c|c|c|c| } 
 \hline
 Metric & Temperature outliers(\%) & Humidity outliers(\%) & Data-Passing Duration (ms) & Duration (ms) &  Billed Duration (ms) & Memory Size (MB) & \$ cost\\ 
  \hline
 $\mu\pm1\times\sigma$ &	5,978 (28.165\%) &	5,706 (26.883\%) & - &	95.47 & - &	32 & \$0.0045\\
  \hline
 $\mu\pm2\times\sigma$ &	446 (2.101\%) &	561 (2.643\%)	 & -  & 102.7 & - &	 30 & \$0.0045\\
  \hline
 $\mu\pm3\times\sigma$ &	6 (0.028\%) &	5 (0.024\%) & - &	85.3 & - &	 31 & \$0.0045\\
 \hline
\end{tabular}
\label{tab:aws_Google_IoT_outlier_detect}
\end{center}
\end{table*}

\begin{table*}
\setlength{\abovecaptionskip}{0pt}
\begin{center}
\scriptsize
\def\tablename{Table}
\caption{\upshape{Performance and cost metrics for performing outlier detection on Raspberry-Pi 4B with Google IoT as the IoT manager.}}
\begin{tabular}{ |c|c|c|p{0.6in}|c|c|c|c| } 
 \hline
 Metric & Temperature outliers (\%) & Humidity outliers (\%) & Data-Passing Duration (ms) & Duration (ms) &  Billed Duration (ms) & Memory Size (MB) & \$ cost\\ 
  \hline
 $\mu\pm1\times\sigma$ &	5,978 (28.165\%) &	5,706 (26.883\%) & - &	326.67 & - &	32 & \$0.0045\\
  \hline
 $\mu\pm2\times\sigma$ &	446 (2.101\%) &	561 (2.643\%)	 & -  & 351.41 & - &	 30 &  \$0.0045\\
  \hline
 $\mu\pm3\times\sigma$ &	6 (0.028\%) &	5 (0.024\%) & - &	291.87 & - &	 31 &  \$0.0045\\
 \hline
\end{tabular}
\label{tab:aws_Google_IoT_pi_outlier_detect}
\end{center}
\end{table*}

\begin{table*}
\setlength{\abovecaptionskip}{0pt}
\begin{center}
\scriptsize
\def\tablename{Table}
\caption{\upshape{Performance and cost metrics for performing outlier detection on Amazon-EC2. We notice that the billed duration is 1 minute (unlike AWS-lambda) as that is the minimum charge per-instance in EC2.}}
\begin{tabular}{ |c|c|c|p{0.6in}|c|c|c|c| } 
 \hline
 Metric & Temperature outliers (\%) & Humidity outliers (\%) & Data-Passing Duration (ms) & Duration (ms) &  Billed Duration (ms) & Memory Size (MB) & \$ cost\\ 
  \hline
 $\mu\pm1\times\sigma$ &	5,978 (28.165\%) &	5,706 (26.883\%) & 404 &	657 &	60000 &	37 & \$0.001417\\
  \hline
 $\mu\pm2\times\sigma$ &	446 (2.101\%) &	561 (2.643\%)	 & 404  & 666 &	60000 &	 37 & \$0.001417\\
  \hline
 $\mu\pm3\times\sigma$ &	6 (0.028\%) &	5 (0.024\%) & 404 &	675 &	60000 &	 37 & \$0.001417\\
 \hline
\end{tabular}
\label{tab:aws_ec2_outlier_detect}
\end{center}
\end{table*}
\bigskip

\begin{table*}
\setlength{\abovecaptionskip}{0pt}
\begin{center}
\scriptsize
\def\tablename{Table}
\caption{\upshape{Performance and cost metrics for performing outlier detection on Google Compute Engine.}}
\begin{tabular}{ |c|c|c|p{0.6in}|c|c|c|c| } 
 \hline
 Metric & Temperature outliers (\%) & Humidity outliers (\%) & Data-Passing Duration (ms) & Duration (ms) &  Billed Duration (ms) & Memory Size (MB) & \$ cost\\ 
  \hline
 $\mu\pm1\times\sigma$ &	5,978 (28.165\%) &	5,706 (26.883\%) & 512 &	770 & 60000 &	32 & \$0.00112\\
  \hline
 $\mu\pm2\times\sigma$ &	446 (2.101\%) &	561 (2.643\%)	 & 512  & 776 & 60000 &	 30 & \$0.00112\\
  \hline
 $\mu\pm3\times\sigma$ &	6 (0.028\%) &	5 (0.024\%) & 512 &	768 & 60000 &	 31 & \$0.00112\\
 \hline
\end{tabular}
\label{tab:Google_compute_outlier_detect}
\end{center}
\end{table*}

\begin{table*}
\setlength{\abovecaptionskip}{0pt}
\begin{center}
\scriptsize
\def\tablename{Table}
\caption{\upshape{Performance and cost metrics for performing outlier detection on Microsoft Azure Virtual Machine.}}
\begin{tabular}{ |c|c|c|p{0.6in}|c|c|c|c| } 
 \hline
 Metric & Temperature outliers (\%) & Humidity outliers (\%) & Data-Passing Duration (ms) & Duration (ms) &  Billed Duration (ms) & Memory Size (MB) & \$ cost\\ 
  \hline
 $\mu\pm1\times\sigma$ &	5,978 (28.165\%) &	5,706 (26.883\%) & 373 &	633 & 1000 &	32 & \$0.000003\\
  \hline
 $\mu\pm2\times\sigma$ &	446 (2.101\%) &	561 (2.643\%)	 & 373  & 645 & 1000 &	 30 & \$0.000003\\
  \hline
 $\mu\pm3\times\sigma$ &	6 (0.028\%) &	5 (0.024\%) & 373 &	463 & 1000 &	 31 & \$0.000003\\
 \hline
\end{tabular}
\label{tab:microsoft_azure_outlier_detect}
\end{center}
\vspace{-5mm}
\end{table*}
\vspace{-10pt}
\subsubsection{Processing on AWS Lambda}
In this section, we evaluate the performance and \$ cost for running our compute-light workload (outlier detection) on the AWS Lambda service. We use the temperature and humidity readings from a single device in the sensor network. Since AWS-Lambda has a number of limitations~\cite{AWS-Lambda-Limits} (such as the maximum timeout of 15 minutes on a single lambda execution), analysis of data points from all 26 devices on our campus network is infeasible for a single Lambda. Therefore, we use a single Lambda per device and report the average performance and \$ cost across all devices. We set the Lambda's max memory to 256 MB. We notice that all other resources (such as CPU compute capacity) is scaled proportionally to the max memory specified. 
We store the data in Amazon S3 and have the Lambda function download directly from S3. We draw several insights. First, the number of outliers decreases with increase in the cut-off threshold [Table~\ref{tab:aws_lambda_outlier_detect}]. However, the runtime is almost identical across the three thresholds. Another major advantage of using AWS Lambda is that it has finer-granularity billing for short-lived jobs vis-\`a-vis other platforms that have a per minute minimum-charge duration (\eg EC2 charges for a 60s minimum duration~\cite{AWS-Ec2-Per-second-billing}). 

\subsubsection{Processing on Emulated device with AWS IoT Greengrass}
Now we show the performance and \$ cost of running our outlier detection workload on an emulated edge device using the AWS Greengrass IoT platform. Moreover, we load the data directly from the container's file-system instead of querying the data from S3. This is performed to show the advantage of running the analysis on the edge, closer to where the data is being collected, hence having a lower execution latency. Table~\ref{tab:aws_green_grass_outlier_detect}
shows the corresponding performance and execution costs for the three cut-off thresholds. AWS Greengrass provides flat pricing per device, so the costs are independent of the execution time. Moreover, we notice the very low execution time compared to AWS Lambda (52--54 msec for emulated edge device with Greengrass vs. $>$ 1 sec for Lambda) without the data-passing overhead.

\subsubsection{Processing on Raspberry-Pi 4B with AWS IoT Greengrass}
Now we show the analysis using a Raspberry-Pi 4B edge device that is connected to AWS Greengrass IoT platform. Again, we load the data directly from the device's file-system rather than querying the data from S3. Table~\ref{tab:aws_greengrass_pi_outlier_detect} shows the corresponding performance and execution costs for the three cut-off thresholds. We notice that the execution times are higher than the execution times for the emulated edge device but less than AWS-Lambda (1 sec for AWS-Lambda $>$ 163--178 msec for Raspberry-Pi with Greengrass $>$ 52--54 msec for emulated edge device with Greengrass) since there is no data-passing overhead.

\subsubsection{Processing on Emulated device with Google IoT Core}
Here we use Google IoT platform and a Docker container that emulates an edge device. We use the same data set as with the previous three platforms and show the performance and cost metrics in Table~\ref{tab:aws_Google_IoT_outlier_detect}. We notice that running outlier detection on an emulated device with Google IoT performs slightly slower than AWS Greengrass. We also notice that in terms of price, Google IoT's pricing model (which is based on the volume of data transfer) shows the highest \$ cost across the 4 platforms. This is because the minimum data size used for billing is 1 MB (which costs \$0.0045). However, Google IoT still provides significantly better performance (\ie, lower latency) compared to AWS-Lambda and AWS-EC2. 

\subsubsection{Processing on Raspberry Pi 4B with Google IoT Core}
Now we evaluate the performance and \$ cost for outlier detection on Raspberry-Pi 4B using the Google IoT platform. Similar to the previous subsection, we notice that running this on the Raspberry Pi with Google IoT performs slightly slower than corresponding platform in AWS.

\subsubsection{Processing on Amazon-EC2}
Here we execute the outlier detection application on an AWS EC2 instance (c5.large). As stated earlier, EC2 has a minimum billing duration of 60 sec~\cite{AWS-Ec2-Per-second-billing}, which makes it more expensive for short-lived jobs compared to AWS Greengrass or AWS-Lambda. Accordingly, we find EC2 to be the most expensive service compared to the other platforms (Table ~\ref{tab:aws_ec2_outlier_detect}). In terms of latency, EC2 also suffers from the data-passing overhead (similar to AWS-Lambda). However, it performs better than AWS-Lambda since it has higher compute capacity.


\subsubsection{Processing on Google Compute Engine}
We show the analysis on Google Compute with an e2-standard machine. As seen with Amazon EC2, the min billed duration is 60s, making it expensive for short-lived jobs. Plus, it suffers from a data-passing overhead.

\subsubsection{Processing on Microsoft Azure Virtual Machine}
Here we show the analysis on Microsoft's Azure Virtual Machine. Unlike Amazon EC2 or Google Compute, the minimum billed duration is 1 sec, which makes it more suitable for short-lived jobs. As such, the cost seen on Table~\ref{tab:microsoft_azure_outlier_detect} is the least compared to all other costs. However, it also suffers from the data-passing overhead, as outlined above.

\subsection{Experiments and Results (Object Detection)}


Here, we show the performance and \$ cost of the object detection workload on various platforms. In the case of processing on the edge device, the video frames are stored on the device for processing. In the case of cloud platforms and other commercial offerings, the video frames are uploaded to Amazon S3 and streamed from there.

We also show results with different number of proposals for Faster RCNN to highlight the advantage of tunability of open source algorithms. This can have an impact on the execution time as shown in Table \ref{tab:rekog_vision_object_detection}. The running time decreases by 57.3\% when approximating aggressively compared to the default parameter value for the number of partitions. It can also have an impact on the mean Average Precision (mAP), as seen in the table. For the same aggressive approximation setting, accuracy decreases by 9\% compared to the default value.  

\begin{table}[ht]
    \begin{center}
        \small
        \def\tablename{Table}
        \caption{\upshape{Performance and cost metrics for running compute-intensive operations, specifically, object detection on Amazon Rekognition, Google Vision, and Microsoft Cognitive Services and an open-source software package, FRCNN.}}
        \label{tab:rekog_vision_object_detection}
        \begin{tabular}{ |p{0.45in}|p{1.0in}|p{0.4in}|p{0.4in}|p{0.4in}|} 
          \hline
         Type & Platform & Accuracy & Frames/ \$ & Time/ frame \\ 
          \hline
          \multirow{1}{*}{\scriptsize Open-source,\small} & FRCNN (100 proposals) on Raspberry-Pi 4B & {\bf 59.11\%}  & - & 23.984 sec\\
          \cline{2-5}
          \multirow{2}{*}{\scriptsize Edge \small} & FRCNN (50 proposals) on Raspberry-Pi 4B & 58.53\% & - & 16.945 sec\\
          \cline{2-5}
          & FRCNN (10 proposals) on Raspberry-Pi 4B & 50.13\% & - & {\bf 10.234 sec}\\
          \hline \hline
          \multirow{1}{*}{\scriptsize Open-source,\small} & FRCNN (100 proposals) on Amazon EC2 & 59.11\% & \textbf{77,266} & 2.318 sec \\
          \cline{2-5}
          \multirow{2}{*}{\scriptsize Cloud \small} & FRCNN (100 proposals) on Google Compute Engine & 59.11\% & 24,666 & {\bf 2.178 sec} \\
          \cline{2-5}
          & FRCNN (100 proposals) on Microsoft Azure Virtual Machine & 59.11\% & 59,306 & 3.02 sec \\
          \hline \hline
          \multirow{1}{*}{\scriptsize Commercial, \small} & Amazon Rekognition & - & {\bf 1000}  & 0.633 sec \\
          \cline{2-5}
          \multirow{2}{*}{\scriptsize Cloud \small} & Google Vision & - & 444 & {\bf 0.471 sec}\\
          \cline{2-5}
          & Microsoft Azure Cognitive Services & - & 500 & 0.488 sec\\
          \hline
        \end{tabular}
    \end{center}
\end{table}


Our first observation for performance is that our custom service is up to 49X slower than Amazon's, Google's and Microsoft's commercial object detection services. However, this is offset by the fact that running a large job with many images can cost more as shown by the lower frames/\$ for the commercial offerings as opposed to running open-source algorithms on the cloud. The decision is left up to the user to evaluate the tradeoff between runtime and cost with the help of benchmarking efforts like \name. This shows the advantage of more evolved commercial services on reducing the latency and providing speedy detection results. However, some drawbacks of commercial offerings are that they are a black box and do not offer tuning knobs that can trade latency for accuracy or price. Furthermore, they do not provide a metric for accuracy, and they do not allow the user to pick the backend on which they run. 

The next observation is that among the commercial services, Google's is less \$ efficient than Amazon's (55\%) and Microsoft's (11\%). However, it is the fastest performer---3\% faster than Microsoft Azure Cognitive Services and 25\% faster than Amazon Rekognition. This can also be seen when running the open-source algorithm on the cloud, where FRCNN on EC2 is the most \$ efficient while FRCNN on Google Compute is the fastest per frame.

\section{Conclusion}
In this paper, we presented \name, the first benchmarking effort of edge computing platforms for different kinds of IoT workloads. We profile Amazon and Google's edge offerings for a compute-light IoT workload (outlier detection on sensor data) and a compute-intensive IoT workload (object detection on streaming video). For the object detection workload, we also use the proprietary Amazon, Google, and Microsoft computer vision offerings and benchmark them against an open source package called Faster R-CNN. Our results show that for compute-light workloads, edge-based services like AWS Greengrass and Google IoT provide the best performance and \$ cost, with AWS Greengrass delivering up to 2X lower latency and up to 1.25X lower cost compared to Google IoT. In contrast, for compute-intensive workloads, the magnitude of tradeoff between latency/execution time and cost is non-trivial. We show that a custom service can be up to 49X slower if run on a slow edge device and up to 6X slower if run on a cloud virtual machine vis-\`a-vis proprietary solutions by Google, Amazon, or Microsoft. We also show how to speed up the open-source solution by approximating aggressively, reducing runtime by 57.3\% at the cost of 9\% drop in accuracy, which highlights the tunability of custom solutions. 

{\small
\setlength{\bibsep}{0.5pt}
\bibliographystyle{acm}
\bibliography{References}
}

\end{document}